# Line Balancing in the Modern Garment Industry

Prof Dr Ray Wai Man Kong[1], Ding Ning[2], Theodore Ho Tin Kong[3]

[1]Adjunct Professor, System Engineering Department, City University of Hong Kong, China
[1]Director, Eagle Nice International (Holding) Ltd., Hong Kong, China
[2]Enginering Doctorate Student, System Engineering Department, City University of Hong Kong, China
[3] Graduated Student, Master of Science in Aeronautical Engineering, Hong Kong University of Science and Technology, Hong Kong
[3] Thermal-acoustic (Mechanical) Design Engineer at Intel Corporation in Toronto, Canada

*Abstract:* This article presents applied research on line balancing within the modern garment industry, focusing on the significant impact of intelligent hanger systems and hanger lines on the stitching process, by Lean Methodology for garment modernization. It explores the application of line balancing in the modern garment industry, focusing on the significant impact of intelligent hanger systems and hanger lines on the stitching process. It aligns with Lean Methodology principles for garment modernization. Without the implementation of line balancing technology, the garment manufacturing process using hanger systems cannot improve output rates. The case study demonstrates that implementing intelligent line balancing in a straightforward practical setup facilitates lean practices combined with a digitalization system and automaton. This approach illustrates how to enhance output and reduce accumulated work in progress.

*Keywords:* Line Balancing, Production Plan, Garment, Automation, Garment Manufacturing, Lean Practice.

## I. INTRODUCTION

The garment industry is known for its labour-intensive tasks, making it one of the most challenging sectors in terms of efficiency. Inefficiencies in workforce utilization often lead to low overall efficiency, hindering productivity and profitability. To address this issue, line balancing techniques have proven to be effective in optimizing operations without incurring additional costs. By matching the output from each operation and calculating operator capacity, line balancing ensures more efficient utilization of resources. This paper aims to provide a systematic approach to line balancing in sewing assembly lines within the garment industry, specifically focusing on the intelligent sewing hanger line balancing method. The study offers valuable guidelines for industries to analyse their systems and improve efficiency, ultimately maximizing output.

## II. GARMENT LINE BALANCING PROBLEM

*Problem of Line Unbalancing in Garment Manufacturing*
Line balancing in the garment industry is the technique of levelling the output of every operation in a garment sewing production line, so the output from the upstream workstation can be optimized to pass through the downstream workstation. There should neither be an accumulation of work between two processes (operation) nor a shortage of workpieces from the upstream workstation (previous work step) between the assembly line and its inter-process. It is important to maintain this balance because in an assembly line output of one process is the input of another.

A production line is not balanced; hence, there would be the following production problems:

- ◆ More Accumulate WIP:
  Some operations can produce more, and some can produce less, which will increase the production line's Work In Progress (WIP).
- ◆ Reduced Efficiency:
  In an imbalanced assembly line, the flow of input and output is uneven. It means that an upstream operation output is a downstream operation input. Because of this reason, some worker will not get loading input as per their capacity of producing output, hence they will be underutilized. In this case, it is to make matters worse more machines and manpower will be allocated to increase production, but efficiency will fall even more.
- ◆ Chaos on the production floor:

Front-line management and workers push themselves to produce more work in process as the chaos of non-bottleneck operations at the imbalanced production line with no results because without improving the line balance all the other efforts will be wasted.

A production line in the garment industry has various layouts of production. The production layouts include the batch production layout without lining up to the production line. Although there is no visual production line, the theoretical calculation can be applied to the line balancing of the process, not limited to a visual conveyor line, one-piece flow line or hanger line. The imbalanced line balance is considered to be the relationship of the manufacturing process between upstream operation and downstream operation. It is not balanced the line balance; hence, the following production problems have existed as shown above points.

### III. LITERATURE REVIEW

Prof Dr Ray Wai Man Kong [1] has mentioned how to shorten the Standard Applied Minutes (SAM) and how to balance the capacity of the machine, machine centre, and work centre in the first level of line balancing of output rate. The increment of individual machines' machinery capacity and assembly capacity cannot improve the whole garment production output and productivity because of line unbalancing problems. The article Lean Methodology for Lean Modernization provides the methodology of how to apply lean technology to work out the future state of value stream mapping (VSM) and goal and find the bottleneck of the garment manufacturing process to enhance the capacity for balancing the whole production process.

Referring to Ocident Bongomin [2], Assembly Line Balancing Problem (ALBP) also known as assembly line design, is a family of combinatorial optimization problems that have been widely studied in literature due to its simplicity and industrial applicability. ALBP is an NP-hard as it subsumes the bin packing problem as a special case. ALBPs arise whenever an assembly line is configured, redesigned, or adjusted.

Published literature shows that the scope of the ALBP in research is indeed quite clear, with well-defined sets of assumptions, parameters, and objective functions. However, these borders are frequently transgressed in real-life situations, in particular for complex assembly line systems like most garment manufacturing. The applied line-balancing problems in garment manufacturing evolved because garment assembly line poses unique balancing problems to those of large body assembly lines such as trucks, buses, aircraft, and machines.

It consists of distributing the total workload for manufacturing any unit of the products to be assembled among the workstations along the line subject to a strict or average cycle time.19 The general principles of line balancing are (1) industrial environments for which the line balancing problems considered are machining, assembly, and disassembly; (2) number of product models: single-model lines, mixed-model lines, multimodal lines; (3) line layout: basic straight line, straight lines with multiple workplaces, U-shaped lines, lines with circular transfer.

The assembly line balancing (ALB) problem has been studied by enterprises for many decades by Gary Yu-Hsin Chen [3]. The ALB model ensures that the staff assignment balances the whole production process to effectively reduce production time or idle time. To meet the ALB, employees' mastery of skills at each task would be considered as an indicator.

However, there are few studies investigating multifunctional (multitasking) workers with multiple levels of skills working at workstations. Our research incorporates the concept of the Toyota Sewing System (TSS) derived from the Toyota Production System (TPS) for the clothing or footwear industry. TSS is credited with less floor space, flexibility and a better working environment. TSS is featured with a U-shaped assembly line and teams of workers making garments on a single-piece flow basis.

Chen *et al*. [4] address a multi-skill project scheduling problem for IT product development. In their research, the project is divided into multiple projects which are completed by a skilled employee. To solve the scheduling problem, they proposed a multi-objective nonlinear mixed integer programming model. Their research takes into consideration employees' skill proficiency at performing tasks, multifunctional employees and cell formation to minimize the production cycle time. Also, adopt another manner to calculate the cycle time different from the previous studies and further consider the workers' skills to reflect the real-world situation. They find that the production time can be effectively reduced with better personnel assignment and a preferred mode of production system. The human factor is an uncertainty to affects the actual cycle time. It is clarified that is the human factor for actual output and driving the real-time dynamic line balancing of garment assembly.

Hoa Nguyen Thi Xuan [4] advised the Applying Genetic Algorithm for Line Balancing Problem in Garment manufacturing and mentioned that Muhammad Babar Ramzan (2019) used a time study approach to balance the line and improve productivity with results in a 36% increase in the machine productivity, reduction of work in process and visibility of the processes also improved. Haile Sime & Prabir Jana (2018) used Arena simulation software to prove the use of simulation techniques in designing and evaluating different alternative production systems from which the one with the best performance can be selected for final implementation. This will help apparel industries to optimize the utilization of their resources through effective line balancing. Markus Proster & Lothar Marz (2015) have shown that dynamic balancing is crucial for high productivity in mixed–model assembly lines to handle the different assembly times of the variants. Common possibilities to treat the resulting capacity peaks are drifting and the allocation of jumpers. A simulation tool was shown that can simulate and visualize these methods and therefore reduce complexity and raise transparency in the planning of assembly lines.

Ghosh and Gagnon (1989) as well as Erel and Sarin (1998) provided detailed reviews on these topics. Configurations of assembly lines for single and multiple products could be divided into three production line types, single–model, mixed–model and multi–model. Single–model assembles only one product, and mixed–model assembles multiple products, whereas a multi-model produces a sequence of batches with intermediate setup operations (Becker & Scholl, 2006).

## IV. METHODOLOGY

### A. Industrial Engineering and Lean Technology to Study the Line Balance of Garment

Referring to Prof Dr Ray Wai Man Kong's article, Lean Methodology for Garment Modernization, industrial engineering and lean study are required to study the whole garment manufacturing process flow. Industrial engineering studies in requirement study play a crucial role in the line balancing of garment manufacturing. An industrial engineer is working for the Here's how it is utilized:

(a) Work Measurement: Industrial engineering study involves conducting time and motion studies to measure the time taken to perform each operation in the sewing assembly line. This data is essential for calculating each operation's cycle time and determining each operator's capacity.

(b) Standardized Work Methods: Industrial engineers analyse the work methods used in garment manufacturing and identify opportunities for improvement. They develop standardized work methods that optimize efficiency and reduce variability in the production process. These standardized methods contribute to effective line balancing.

(c) Garment Process Analysis: Industrial engineers analyze the entire garment manufacturing process, from receiving raw materials to the final product's shipment. The garment sewing process bottlenecks, inefficiencies, and areas of improvement can be identified by Value Stream Mapping. By understanding the process flow, the future state of VSM can identify opportunities for line balancing and optimize the sequence of operations.

(d) Capacity and Manpower Resource Allocation: Industrial engineers assess the garment sewing workforce and equipment available in the garment manufacturing facility. They determine the number of operators required for each operation based on the calculated cycle time and operator capacity. This helps in allocating resources effectively and achieving a balanced line.

(e) Layout Design: Industrial engineers consider the layout of the sewing assembly line and its impact on efficiency. They analyze the flow of materials, equipment placement, and operator movement. By optimizing the layout including (1) batch layout, (2) one-piece line layout, (3) conveyor line layout and (4) intelligent line layout, besides the batch layout, other 3 production line layouts for the sewing process can minimize unnecessary movement, reduce transportation time, and improve overall line balancing.

(f) Ergonomics and Workplace Design: Industrial engineers consider ergonomics principles to design workstations that promote worker safety, comfort, and productivity. They ensure that the layout and design of workstations support efficient movement and minimize fatigue. This contributes to improved line balancing by enhancing operator performance.

(g) Continuous Improvement: Industrial engineering study emphasizes continuous improvement in line balancing as referred to the Lean Methodology for Garment Modernization. Industrial engineers monitor the line's performance, collect data, and analyze it to identify areas for further optimization. They implement changes, conduct follow-up studies, and refine the line-balancing process to achieve higher efficiency and productivity.

By utilizing industrial engineering study in line balancing of garment manufacturing, companies can optimize their production processes, reduce lead times, improve resource utilization, and enhance overall efficiency. This results in increased productivity, cost savings, and improved customer satisfaction.

*B. General Garment Manufacturing Process*

Before the line balancing for the sewing process, the structure of the garment is separated into two major manufacturing processes. The first one is the part sewing which seems that individual parts sewing. Garment. components are the basic sections of garments including top fronts, top backs, bottom fronts, bottom backs, sleeves, collars/neckline treatments, cuffs/sleeve treatments, plackets, pockets, and waistline treatments. A few processes are involved in the buttoning, ironing and other equipment for elastic sewing on garment parts which is counted on the part assembly or part sewing process. In the garment factory, it is called the sub-assembly process. Template sewing is one of the automated processes in the automation of part assembly. The second one is the final main assembly which gets the part assembly to combine to the finished garment. After the garment has been finished with all related main assembly processes, the last operation is trimming, ironing, packing to the polybag and then packing to the carton box.

*C. Conveyor Line of Sewing Process for Line Balance of Garment*

In the traditional batch production layout, the sub-assembly process and main assembly process are located on the same production floor. There is a typical ALBP that can be applied to various mathematic methods to optimize the line balancing, but the travel time is counted for the bundle batch for transportation between one workstation (sewing station) to another workstation. Bundle batch assembly is not easy to handle on transportation and not easy to visualize any overstock of work in progress at production floor.

According to Prabir Jana [5], the sewing workshop layout has set up two rows of sewing machines and the centre table in between is a common view in almost all factories. The origin of the centre table is not known but it is probably meant to solve two problems together: to provide a cover to electrical wiring and to act as a material storage/transfer facility. During the mid-'80s, there were some factories with conveyors (in place of the centre table) in between two rows of machines for the sub-process of garment sewing and main garment sewing as the Batch sewing line for the garment assembly as shown in Fig 1. It is difficult to make the line balance in actual garment production. The front-line supervisor uses the visual way to monitor any overstock of production in between sewing stations.

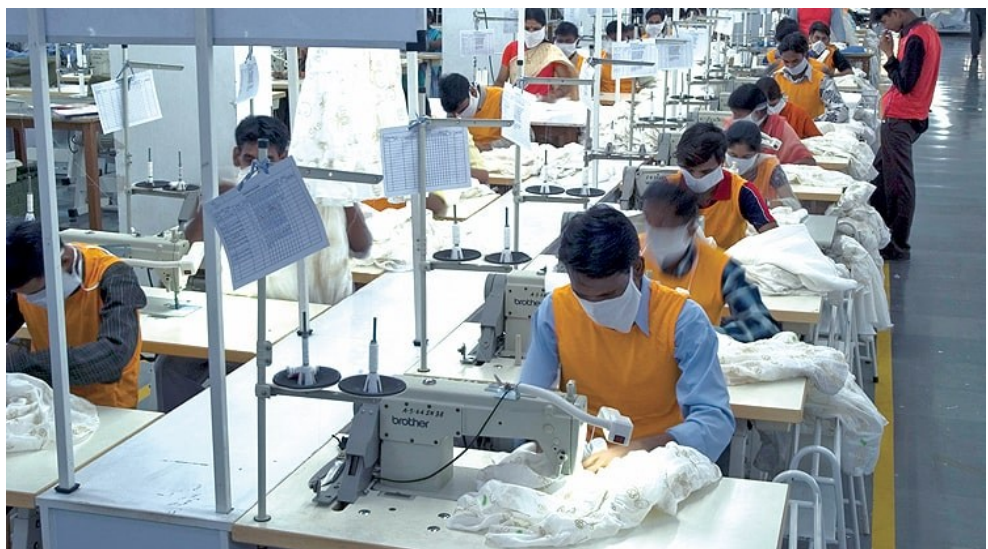

Figure 1 Batch sewing line for the garment assembly

In the conveyor line and layout in the main assembly, there is a modernization method and way to reduce the travel time between workstations and improve visual manufacturing and front-line control as shown the Figure 2. The conveyor line and conveyor line layouts have the benefit of line balance on the main assembly output and enhanced efficiency. The Hanger conveyor layout is applied to intelligent manufacturing for garments. Because it does not use the progressive bundle concept, this style of layout eliminates the previous Work-in-Progress. allows all the materials for a specific garment to be transferred as a unit to any workstation's sewing machine. When an operation at one workstation is completed, the operator should hang the garment to the hanger and then press a button to confirm the finished unit work by faster clipping, so the hanger system can deliver the work-in-progress unit of the garment to the next workstation either mechanically or automatically. It can reduce material handling time. Such a system's layout must be continuous, with no gaps in between. The materials flow through the layout in a loop shape. The hanger line is required to construct the hanger system and equipment. The system is modernized to set up the control device to move the hanger between workstations and provide the just-in-time information to the manufacturing system. The line balancing for the hanger line can be optimised to increase production efficiency by increasing the through-put time based on increased the capacity of the bottleneck workstations in the process as the Lean Methodology for Garment Modernization that Prof Dr Ray WM Kong mentioned [1].

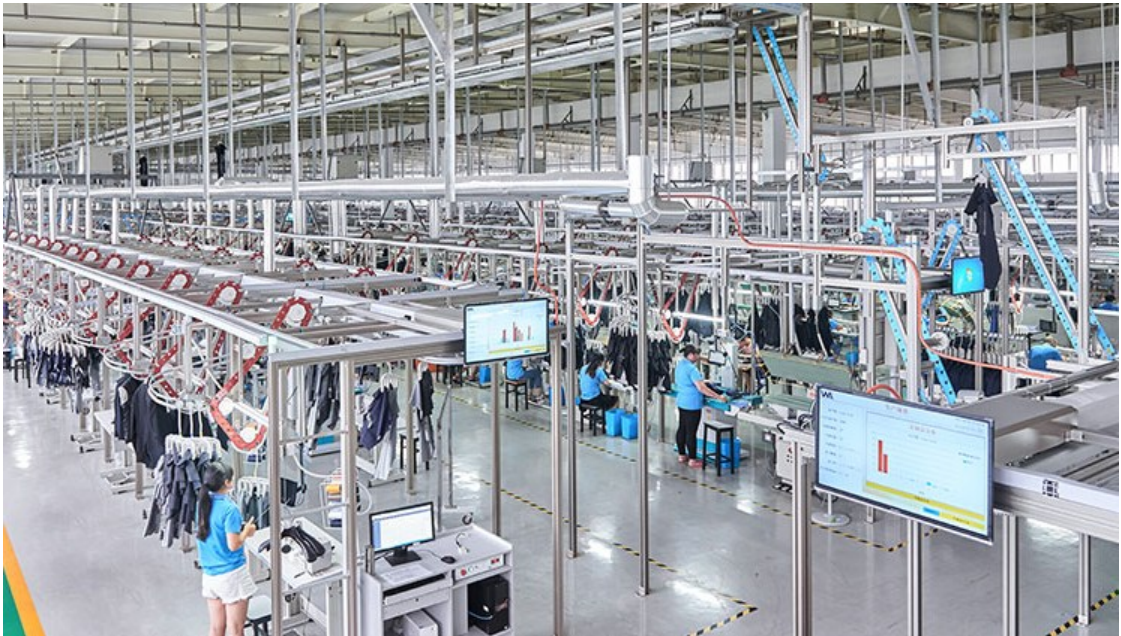

Figure 2 Intelligent Hanger Line and System from INA Intelligent Technology (Zhejiang)

## V. MIXED LAYOUT FOR INTELLIGENT HANGER CONVEYOR LINE AND BATCH SEWING WORKSTATION

Garment manufacturing is the most comprehensive process for the line layout design. The prerequisite line balance is required to set up the appropriate line layout for garment manufacture based on various types of garment categories and garment styles. The garment category clarifies the various types of garments: Polo shirts, Dresses, Jeans, Jackets, Pants, Leggings, sportswear, swimwear and others. The garment styles include various garment constructions: grommet drawcord, buttonhole drawcord, inseam gusset, banded hem, banded hem, bound hem exposed trim, elastic of front, zipper of pocket, on-seam pocket and others.

The problem with the fixed facility of the intelligent hanger line is that it cannot be optimized for both part assembly and man flow assembly. The part assembly involves a short cycle time and participating sewing workmanship skills. Optimization and high efficiency are required to reduce the setup time for the part garment assembly in the batch production layout. The operator should continue to produce the same sewing process in the part assembly repeatably. An operator does not change threads, sewing needles and pulling folders if required. The skilled operator can get the benefit of division of work with less change of style, garment construction, fabric piece and upstream sub-assembly work pieces.

Batch production is a method of producing sub-assembled garments in batches, or groups, of a fixed quantity and quality. Batch production plant layout design involves arranging the equipment and facilities in a way that allows each batch to go through a series of operations in a sequential order. The batch production can count as the individual operation of part sewing by the bundle. Bundling is an essential sub-process in the fabric-cutting department of garment manufacturing at the beginning of processes. It involves arranging the cut garment fabric piece from the cut stacks after layer cutting and tying them together in bundles. Each bundle typically contains a specific number of garment fabric pieces for part assembly as a bundle batch production model. A bundle batch production model is required for the sub-assembly of the garment for batch production to get the benefit of high efficiency and output based on the reduction of waiting time, machinery setup time and changeover time on changing thread for the garment.

The mixed layout for the Intelligent Hanger Conveyor Line and Batch Sewing Workstation can be created to build the subassembly part sewing in the batch production layout by batch and main consequent sewing in the hanger line for main garment assembly. The hanger line is defined by one-piece production flow, not batch production. One-piece flow is the continuous process flow, or single-piece process flow is a discrete production method used in lean manufacturing. The article Lean Methodology for Garment Modernization has provided the methodology and process of how to work out the overall lean practice in garment modernization [1].

Mixing line layouts for batch production and flow line production can be an effective strategy to optimize manufacturing processes, especially in environments where product variety and volume fluctuate. Here's how you can achieve a mixed line layout that incorporates elements of both batch and flow line production:

1. Understanding the Characteristics:
   - Batch Production: Involves producing goods in groups or batches. It is flexible and allows for variations in product types but may lead to longer lead times and higher work-in-progress (WIP) inventory.
   - Flow Line Production: Involves a continuous flow of production with a fixed sequence of operations. It is efficient for high-volume production of standardized products but lacks flexibility.

2. Identify Product Families:
   - Group products based on similarities in processing requirements, production volume, and demand patterns. This will help in determining which products can be produced in batches and which can be produced in a flow line.

3. Designing the Layout:
   - Cellular Manufacturing: Create manufacturing cells that combine batch and flow production. Each cell can be designed to handle a specific product family, allowing for batch production within the cell while maintaining a flow line for certain operations.
   - The hanger line in cellular manufacturing is a one-piece manufacturing flow which can be flexible to amend the workstation, template sewing machine, single needle machine, double needles sewing machine, overlock sewing machine and ironing station for various assemblies in the main process flow.
   - Flexible Workstations: Design workstations that can accommodate both batch and flow production. For example, a workstation can be equipped with tools and equipment that allow operators to switch between batch processing and flow processing as needed.
   - The hanger line is equipped for the main garment assembly. The part assembly as the stitching process and special operation (Ig. Buttoning, embroidery …etc.) are set in the batch production layout because of individual operations and do not rely on the upstream operations. The batch production layout and batch production can shorten the cycle time because of the reduction of changeover time and speed up output by down the skill of operation in the repeatable works.
   - U-shaped layout: Consider a U-shaped layout that allows for easy movement of materials and operators. This layout can facilitate both batch and flow production by enabling quick transitions between different production modes.
   - The hanger line in a long U-shaped layout means that the hanger line can be equipped for the U shape.

In Figure 3, the Batch Production and Intelligent Hanger Line Flow Production Layout shows the hanger line that set up the U-shaped layout.

4. Implementing Hybrid Production Strategies:
   - Mixed-Model Production: Implement a mixed-model production approach where different styles are produced in a single line. This can be achieved by scheduling production runs based on demand forecasts, allowing for both batch and flow production within the same garment line.
   - It can cause the potential risk of wrong route and using the wrong thread. For lean manufacturing, the lean practice has a concept to prevent any mistake and failure of garment manufacturing. I have not suggested working to work out mixed-model production although there is a workable and operational solution.
   - Kanban System: Use a Kanban system to manage inventory and production flow. This system can help control the flow of materials and products between batch and flow operations, ensuring that the right amount of product is produced at the right time.

- Separated the cutting fabric piece in the batch layout, not mix to the stitching production line layout, the cutting piece stock can be kept on the bundle up for the panel cutting piece stockroom as a modern automated warehouse. The cut piece can be provided to hanger line or part assemble line based on the Kanban.

5. Balancing Workloads:
    - Analysis of the existing sewing assembly line is required to make any change in process and setup.
    - Time study and calculation of cycle time for each operation is required to provide the standard time in each operation.
    - Determination of operator capacity and workload distribution is executed to plan the available manhours a day and the number of work days a week for the production plan.
    - Balancing the line workload by selecting the appropriate equipment, machine and tools for optimizing resource allocation.
    - Monitoring and continuous improvement of the balanced line :
    Analyze the workload at each workstation to ensure that operators are not overburdened or underutilized. Adjust the number of operators and the sequence of operations to achieve a balanced workload that accommodates both batch and flow production. The next section is related to the mathematic model how to calculate the required of workload and available capacity to make a balance of workload.

6. Utilizing Technology:
    - Implement technology such as automated guided vehicles (AGVs) or hanger lines with its systems that can support both batch and flow production. These technologies can help transport materials and products efficiently between different production areas. AGV can carry the holder of fabric pieces from the cutting room to the stitching room.

7. Continuous Improvement:
    - Regularly review and analyze production performance to identify areas for improvement. Gather feedback from operators and use data analytics to optimize the mixed layout continually. This may involve adjusting the layout, reconfiguring workstations, or changing production schedules.
    - The intelligent hanger system can get the on-time input and output from the workstation of the hanger line. It can provide the on-time measurement and output rate to the front-line supervisor. The front-line supervisor can amend the number of workstations by splitting from one garment operation to two stations. It means that more than one workstation produces the same garment works as reacting on the line balance.

8. Training and Cross-Training:
    - Train operators to be versatile and capable of working in both batch and flow production environments. Cross-training employees can enhance flexibility and responsiveness to changing production needs.

By thoughtfully integrating elements of batch and flow hanger line production, garment manufacturers can create a multiple-line layout that maximizes efficiency, flexibility, and responsiveness to customer demands referred to as the Lean Methodology for Garment Modernization [1]. This approach allows for both production methods' benefits while minimising their drawbacks, ultimately leading to improved productivity and customer satisfaction.

## VI. LINE BALANCING MODEL ANALYSIS

The main objective of balancing the line is to produce the same expected number of work-in-progress output rates for every process of an assembly line. There should neither be an accumulation of work between two processes nor an absence of work between the processes. It is important to maintain this balance because in a garment assembly line output of one process is the input of another process. The target of the upstream operation rate should be the same or slightly greater than the downstream operation rates; hence, the is no shortage of supply WIP to the next operation as reduces the idle time of the next operation. The target of the downstream operation rate should be the same or slightly greater than the upstream operation rates; hence, the is no overstock of WIP between the previous operation and the next operation.

A cutting process is the first main manufacturing process after the incoming quality inspection. The batch production layout is the most common to maximize the output of fabric pieces because modern high-speed cutting workbenches can maximize the output through marker design for optimization of fabric for cutting process. The technical detail of the cutting operation is not covered in this section.

According to the typical sewing process chart from Prasanta Sarkar [6], the stitching process means a sewing process. Based on the process chart, the main flow has been defined as the major continuous flow.  It is appropriate to set it up in the hanger line.  Other part assembly and part sewing can be separated to the part assembly in the batch production layout which can be separated to the front of the hanger line as reduced the travel time of bundle work in progress based on the lean practice.

The line balance in the hanger line can apply a similar mathematical model of line balance for the production flow line.  Referring to the line balancing method from Prof Dr Ray WM Kong [1], the bottleneck operation should be found to try to increase the capacity until the bottleneck operation shifts to the next bottleneck operation or accept the throughput rate to meet the future state of value stream mapping.

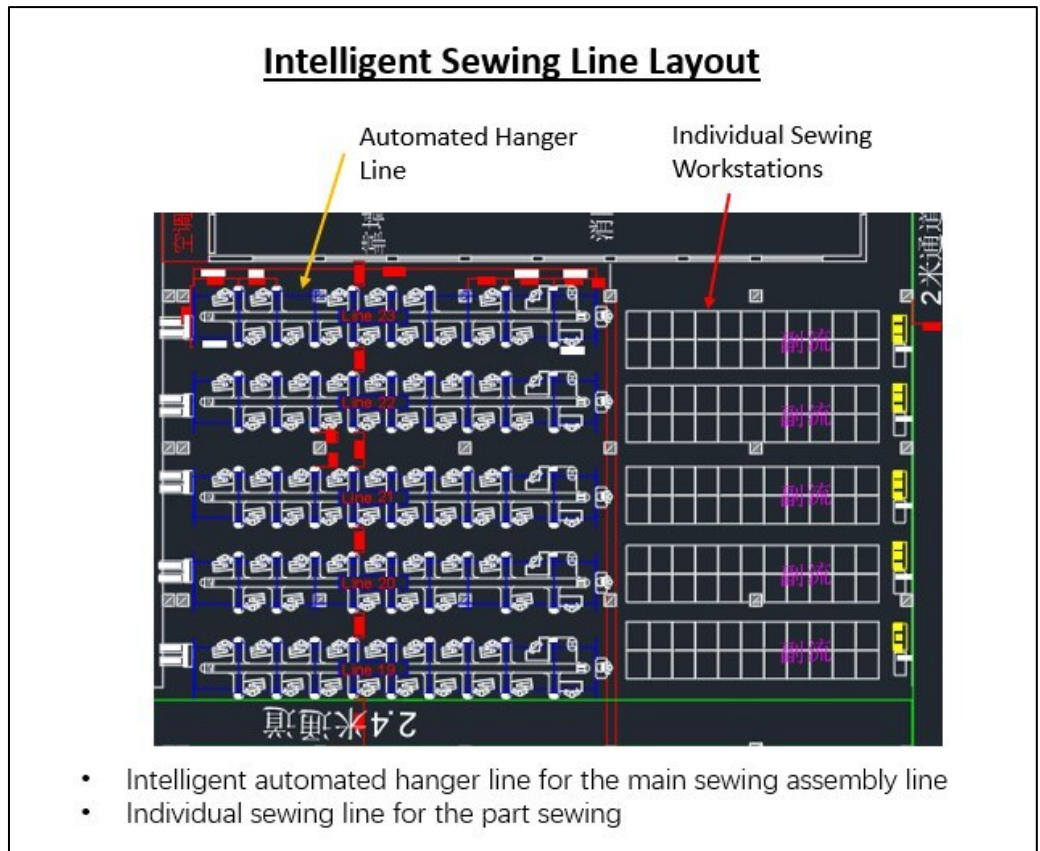

Figure 3 Batch Production and Intelligent Hanger Line Flow Production Layout

Balancing in the stage of a single model finds a locally optimized solution in iteration. The objective of the stage is to find a solution(s) with a specified number of stations with a minimum cycle time.  Solutions are considered locally optimized as the principal objective is to find a solution which will define a smooth production by minimizing the objective function of assembly workstation balance from Waldemar Grzechca [7]. The concept of ALBP, where the aim is to optimize the number of workstations with a predefined fixed cycle time is utilized in the formulation. The fixed cycle time is considered the solution lower bound, $CT_{min}$ for finding desired station numbers, with minimum cycle time as the output rate of hanger line.

The minimize of cycle time is shown in below formula:

$$CT_{min} = \max \left[ \frac{1}{S}\sum_{i=1}^{n} t_i, \max t_i \right] \quad (1)$$

$$minCT = \max \left[ \frac{1}{S}\sum_{i=1}^{n} t_i, \max t_i \right] \quad (2)$$

subject to
$$S \leq 32 \quad (3)$$

Where $t_i$ is the $i_{th}$ task time and $S$ is the desired number of workstations which means that sewing machine and other machines.

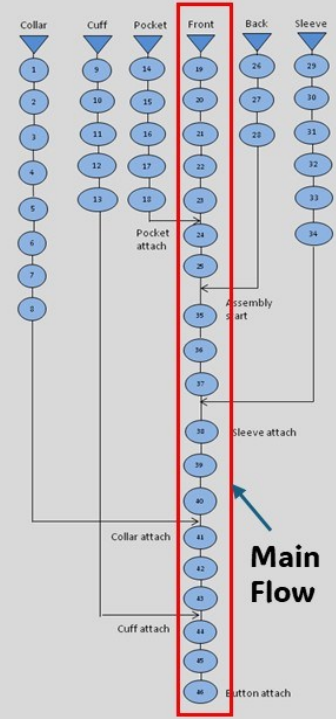

Figure 4 Batch Production and Intelligent Hanger Line Flow Production Layout

**TABLE I: Shirt Garment Assembly for Hanger Line (Before line balancing)**

| Task | Description | Cycle Time (sec/pc) | Number of workstation | Remark |
|---|---|---|---|---|
| 19 | Mark Front for Pocket Position | 30 | 1 | |
| 20 | Form Buttonhole plackets | 40 | 1 | |
| 21 | Crease B/H Placket (Single Fold) | 60 | 1 | (5th High CT) |
| 22 | Top stitch B/H placket | 40 | 1 | |
| 23 | Sew Button Placket | 25 | 1 | |
| 24 | Attach pocket | 20 | 1 | |
| 25 | Sewlabel at placket | 50 | 1 | (6th High CT) |
| 35 | Set front & back & mark neck for collar | 60 | 1 | (5th High CT) |
| 36 | Shoulder attach | 60 | 1 | (5th High CT) |
| 37 | Shoulder top stitch | 120 | 1 | (1st Highest CT Bottleneck Operation) |
| 38 | Sleeve Attach | 40 | 1 | |
| 39 | Top stitch armhole | 80 | 1 | (3rd Highest CT) |
| 40 | Side Seam | 110 | 1 | (2nd Highest CT) |
| 41 | Collar Attach | 30 | 1 | |
| 42 | Collar Close & Insert Label | 60 | 1 | (5th High CT) |
| 43 | Cuff Attach & Close | 80 | 1 | (3rd High CT) |
| 44 | Bottom Hem | 80 | 1 | (3rd High CT) |
| 45 | Button Hold - Front Placket & Collar | 70 | 1 | (4th High CT) |
| 46 | Button Attach (Last Operation for finishing process) | 35 | 1 | |

| | | | | Because of the bottleneck operation |
|---|---|---|---|---|
| Hanger line output rate | | 120 | sec/pc | |

The main continuous operation is set into the hanger line in the main production flow layout as referred to in Fig. 4. The integer linear programming can resolve the line balance of the hanger line as **Table I** subject to the constraint of 32 seats of hangers which maximizes the output rate as simple iteration as shown as **Table II**. In the case study, the simple mechanism is to seek the bottleneck workstation as the high cycle time to increase the workstation. The number of workstations should be the integer number because of reality without mixing for various garment styles. The output rate hanger line was increased from 120sec/pc to 40sec/pc as 3 times enhancement of output.

**TABLE II: Shirt Garment Assembly for Hanger Line (After line balancing – 32 seats)**

| Task | Description | Cycle Time (sec/pc) | Number of workstations | Cycle Time after counted workstation (sec/pc) | Remark |
|---|---|---|---|---|---|
| 19 | Mark Front for Pocket Position | 30 | 1 | 30 | |
| 20 | Form Buttonhole plackets | 40 | 1 | 40 | |
| 21 | Crease B/H Placket (Single Fold) | 60 | 2 | 30 | |
| 22 | Top stitch B/H placket | 40 | 1 | 40 | |
| 23 | Sew Button Placket | 25 | 1 | 25 | |
| 24 | Attach pocket | 20 | 1 | 20 | |
| 25 | Sew label at placket | 50 | 2 | 25 | |
| 35 | Set front & back & mark neck for collar | 60 | 2 | 30 | |
| 36 | Shoulder attach | 60 | 2 | 30 | |
| 37 | Shoulder top stitch | 120 | 3 | 40 | |
| 38 | Sleeve Attach | 40 | 1 | 40 | |
| 39 | Top stitch armhole | 80 | 2 | 40 | |
| 40 | Side Seam | 110 | 3 | 36.7 | |
| 41 | Collar Attach | 30 | 1 | 30 | |
| 42 | Collar Close & Insert Label | 60 | 2 | 30 | |
| 43 | Cuff Attach & Close | 80 | 2 | 40 | |
| 44 | Bottom Hem | 80 | 2 | 40 | |
| 45 | Button Hold - Front Placket & Collar | 70 | 2 | 35 | |
| 46 | Button Attach (Last Operation for finishing process) | 35 | 1 | 35 | |
| | | Total: | 32 Seats | 40 | (Output rate) |

The average workstation time (s) is simply the total work content ($\sum t$) divided by the actual number of stations ($\sum s$).

$$CT_{min} = \max\left[\frac{1}{s}\sum_{i=1}^{19} t_i, \max t_i\right]$$

Subject to

$S <= 32$ (because of fixed 32 seats in the hanger line)

For the case study in factory X's hanger line, the cycle time before line balance is 120sec/pc. The total work content is 1090sec for 19 workstations (19 workers). It means that 19 workers provide 30pc/hr. [(3600sec / (120sec/pc)). After the line balancing technology is applied to the hanger line, the cycle time is improved to 40sec/pc. The total work content is 1090sec for 32 workstations (32 workers). It means that 32 workers provide 90pc/hr. [(3600sec / (40sec/pc)).

Unit Per Person Hour (UPPH) in *i* number of line balancing iterations,

$$UPPH_i = \frac{Output_i}{H_i} \quad (4)$$

For the case study, the unit per person work (UPPH$_i$) is 0 for the original state before line balancing.

$$UPPH_0 = \frac{30}{19} = 1.57 \text{pcs/man-hour}$$

After the line balancing technique was applied to the hanger line, the UPPH increased from 1.57 pcs/man-hour to 2.81 pcs/man-hour.

$$UPPH_1 = \frac{90}{32} = 2.81 \text{pcs/man-hour}$$

The improvement percentage of line balancing output is compared between the original plan and the new plan. The improvement percentage ($Eff\ improvement$) of line balancing's UPPH is shown following:

$$Eff\ improvement = \frac{(UPPH_1 - UPPH_0)}{UPPH_0} = 78.98\% \text{ (positive of increment of UPPH)}$$

Referring to the Optimization Model for Assembly Line Balancing Problem with Uncertain Cycle Time from Yong Cao et al, the cycle time is not a fixed figure which means it affects the line balancing of the hanger line due to the human factor. Assume that *CT0* is the nominal cycle time and D is the maximum change in the cycle time. The cycle time interval can be obtained as follows:

$$CT \in [CT_0 - D, CT_0 + D] \quad (5)$$

Subject to

$$CT \leq CT_0 + \alpha D \quad (6)$$
$$CT \geq CT_0 - \alpha D \quad (7)$$

Where $\alpha \in (0, 1]$. In the design phase of an assembly line, the production engineer can choose a reasonable value of $\alpha$ to cope with uncertainty in the cycle time.

**TABLE III: Shirt Garment Assembly for Hanger Line with Deviation Factor**

| Task | Description | Cycle Time (sec/pc) | Number of workstation | Cycle Time after counted workstation (sec/pc) | Deviation of Cycle Time +αD | Deviation of Cycle Time -αD |
|---|---|---|---|---|---|---|
| 19 | Mark Front for Pocket Position | 30 | 1 | 30 | 32 | 29 |
| 20 | Form Buttonhole plackets | 40 | 1 | 40 | 42 | 38 |
| 21 | Crease B/H Placket (Single Fold) | 60 | 2 | 30 | 32 | 29 |
| 22 | Top stitch B/H placket | 40 | 1 | 40 | 42 | 38 |
| 23 | Sew Button Placket | 25 | 1 | 25 | 27 | 24 |
| 24 | Attach pocket | 20 | 1 | 20 | 21 | 19 |
| 25 | Sewlabel at placket | 50 | 2 | 25 | 27 | 24 |
| 35 | Set front & back & mark neck for collar | 60 | 2 | 30 | 32 | 29 |
| 36 | Shoulder attach | 60 | 2 | 30 | 32 | 29 |
| 37 | Shoulder top stitch | 120 | 3 | 40 | 42 | 38 |
| 38 | Sleeve Attach | 40 | 1 | 40 | 42 | 38 |
| 39 | Top stitch armhole | 80 | 2 | 40 | 42 | 38 |
| 40 | Side Seam | 110 | 3 | 36.7 | 39 | 35 |
| 41 | Collar Attach | 30 | 1 | 30 | 32 | 29 |
| 42 | Collar Close & Insert Label | 60 | 2 | 30 | 32 | 29 |
| 43 | Cuff Attach & Close | 80 | 2 | 40 | 42 | 38 |
| 44 | Bottom Hem | 80 | 2 | 40 | 42 | 38 |
| 45 | Button Hold - Front Placket & Collar | 70 | 2 | 35 | 37 | 34 |

| 46 | Button Attach (Last Operation for finishing process) | 35 | 1 | 35 | 37 | 34 |
| | Maximize Cycle Time as thought-put time (sec) | | | 40 | 42 | 38 |

According to the case study, **Table III** shows shown the cycle time should be estimated to the deviation, $\pm\alpha D$ to make the impact to the UPPH of line balancing which calculates the line balance near the reality of the actual result.

After the line balancing technique was applied to the hanger line with positive deviation $+\alpha D$ the UPPH before line balancing from 1.57 pcs/man-hour to 2.66 pcs/man-hour. After the line balancing technique was applied to the hanger line with positive deviation $-\alpha D$ the UPPH before line balancing from 1.57 pcs/man-hour to 2.96 pcs/man-hour.

$$UPPH_{max} = \frac{95}{32} = 2.96 \text{pcs/man-hour}$$

$$UPPH_{min} = \frac{85}{32} = 2.66 \text{pcs/man-hour}$$

$$Eff\ improvement|_{max} = \frac{(UPPH_{max} - UPPH_0)}{UPPH_0} = 89\% \text{ (positive of increment of UPPH)}$$

$$Eff\ improvement|_{min} = \frac{(UPPH_{min} - UPPH_0)}{UPPH_0} = 69\% \text{ (positive of increment of UPPH)}$$

The graph chart of cycle time analysis shows the variance level of uncertainty α was assumed to vary within a certain range. The α level determined the cycle time interval. Figure 5 provides three results of cycle time, namely, regular, best, and worst. It should be noted that the robust results considered the uncertainty in cycle time, expressed as an interval, the regular results considered the normal cycle time, and the line improvement best results were calculated to the above result.

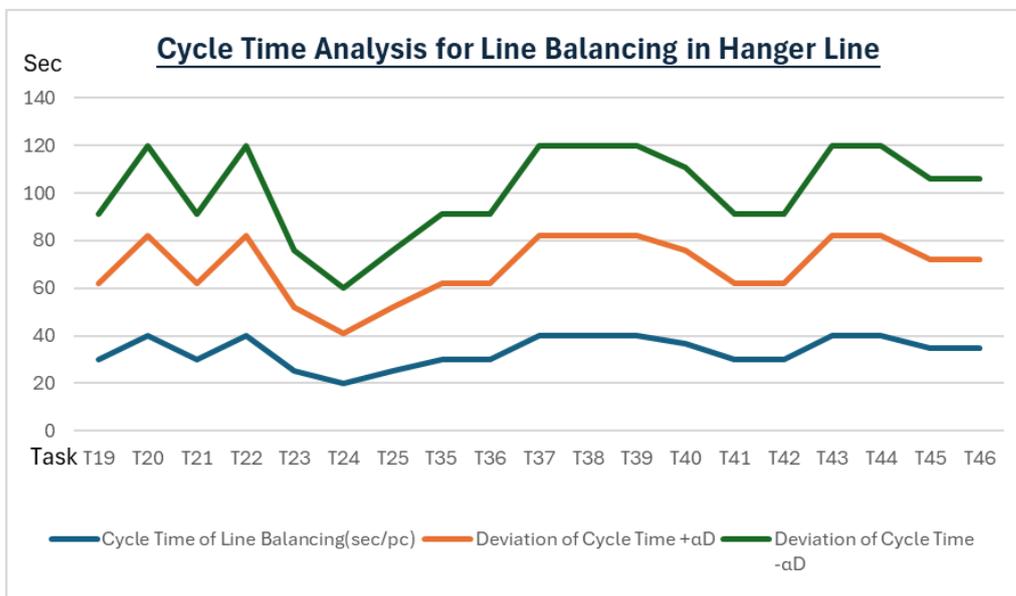

Figure 5  Cycle Time Analysis for Line Balancing in Hanger Line

## IV. CONCLUSION

In conclusion, effective line balancing in garment assembly operations is essential for resolving the issue of excessive work-in-process (WIP) inventory and improving the production output and efficiency, which often arises from an unbalanced production line. By systematically analyzing and optimizing the distribution of tasks among operators, organizations and factory planners can achieve a more synchronized workflow that minimizes bottlenecks and reduces idle time. This not only leads to a smoother production process but also significantly decreases the accumulation of WIP, thereby lowering storage costs and enhancing overall operational efficiency. Ultimately, implementing line-balancing techniques fosters a leaner manufacturing environment, improves responsiveness to market demands, and contributes to higher levels of productivity and profitability in the garment industry.

Furthermore, garment automation and intelligent manufacturing are the most important factors in increasing productivity and output. The development of a research study in the Design and Experimental Study of Vacuum Suction Grabbing Technology to Grasp Fabric Piece can provide the new innovative technology and theory to develop the vacuum gripper for automation from Prof Dr Ray WM Kong et al [9]. Design a New Pulling Gear for the Automated Pant Bottom Hem Sewing Machine from ProfDr Ray WM Kong et al [10] can enhance the production rate of the hem sewing machines shown above case study. Automation is the best opportunity to enhance garment manufacturing and industry.

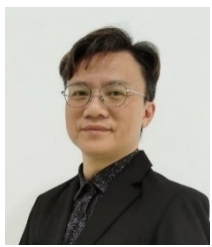

**Ray Wai Man Kong** (Senior Member, IEEE, member of IET, MIET, MPAA) Hong Kong, China. He received a Bachelor of General Study degree from the Open University of Hong Kong, Hong Kong in 1995. He received an MSc degree in Automation Systems and Engineering and an Engineering Doctorate from the City University of Hong Kong, Hong Kong in 1998 and 2008 respectively.

From 2005 to 2013, he was the operations director with Automated Manufacturing Limited, Hong Kong. From 2020 to 2021, he was the Chief Operating Officer (COO) of Wah Ming Optical Manufactory Ltd, Hong Kong. He is a modernization director with Eagle Nice (International) Holdings Limited, Hong Kong. He holds an appointment, as an Adjunct Professor of the System Engineering Department at the City University of Hong Kong, Hong Kong. He published more articles in international journals in robotic technology, gripper, automation and intelligent manufacturing. His research interests focus on intelligent manufacturing, automation, maglev technology, robotics, mechanical engineering, electronics, and system engineering for industrial factories.

Prof. Dr. Kong Wai Man, Ray is Vice President of CityU Engineering Doctorate Society, Hong Kong and a chairman of the Intelligent Manufacturing Committee of the Doctors Think Tank Academy, Hong Kong.  He has published more intellectual properties and patents in China.

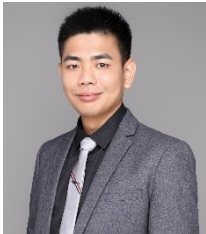

**James Ding Ning**（IEEE member, Engineering Doctorate Student）Hong Kong, China.  He received a Bachelor of Electronic and Information Engineering degree from Shenzhen University, Guangdong Province, China in 2006.  He received an MSc degree in Electronic Engineering from the University of Sheffield, United Kingdom in 2007 with upper second honours. He is a part-time Engineering Doctorate Student of the System Engineering Department at the City University of Hong Kong.

Mr. Ning has more than 16 years of industry experience focused on semiconductor and intelligence instrument design. He is the first author of more than 10 patents. He is currently working at Huawei Investment Limited Corporation, Hong Kong as a Senior Engineer. His research interest is focused on computer architecture design, CPU Design, integrated circuit design, communication standard protocol and industry management. He has extensive work experience in both Shenzhen and Hong Kong, with his professional footprint spanning across the Greater Bay Area.

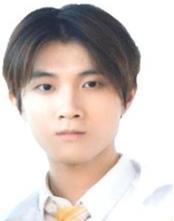

**Theodore Ho Tin Kong** (MIEAust, Engineers Australia and MIEEE) received his Bachelor of Engineering (Honours) in mechanical and aerospace engineering from The University of Adelaide, Australia, in 2018. He then earned a Master of Science in aeronautical engineering (mechanical) from HKUST - Hong Kong University of Science and Technology, Hong Kong, in 2019.

He began his career as a Thermal (Mechanical) Engineer at ASM Pacific Technology Limited in Hong Kong, where he worked from 2019 to 2022. Currently, he is a Thermal-Acoustic (Mechanical) Design Engineer at Intel Corporation in Toronto, Canada. His research interests include mechanical design, thermal management and heat transfer, and acoustic and flow performance optimization. He is proficient in FEA, CFD, thermal simulation, and analysis, and has experience in designing machines from module to heavy mechanical level design.